\pdfoutput=1

\documentclass[journal]{IEEEtran}
\usepackage{graphicx}
\ifCLASSINFOpdf
\else
\fi
\usepackage{amsmath}

\usepackage[caption=false,font=footnotesize]{subfig}

\usepackage[numbers]{natbib}

\begin{document}
	%

	
	\title{Anomaly Detection using Deep Learning based Image Completion}
	
	
	%
	%
	%
	
	\author{{M. Haselmann,
		D.P. Gruber,
		P. Tabatabai}
		\thanks{\copyright 2018 IEEE. Personal use of this material is permitted. Permission from IEEE must be obtained for all other uses, in any current or future media, including reprinting/republishing this material for advertising or promotional purposes, creating new collective works, for resale or redistribution to servers or lists, or reuse of any copyrighted component of this work in other works.}
		\thanks{The authors are with the Polymer Competence Center Leoben GmbH, Roseggerstrasse 12, 8700-Leoben, Austria. E-mail: \{matthias.haselmann, \newline dieter.gruber, paul.tabatabai\}@pccl.at.}
		
	}

	%
	%

	
	\markboth{Accepted for publication by IEEE, 17th International Conference on Machine Learning and Applications (ICMLA), 2018}{}%
	%



	\maketitle
	
	\begin{abstract}
		Automated surface inspection is an important task in many manufacturing industries and often requires machine learning driven solutions. Supervised approaches, however, can be challenging, since it is often difficult to obtain large amounts of labeled training data.
		In this work, we instead perform one-class unsupervised learning on fault-free samples by training a deep convolutional neural network to complete images whose center regions are cut out. Since the network is trained exclusively on fault-free data, it completes the image patches with a fault-free version of the missing image region. The pixel-wise reconstruction error within the cut out region is an anomaly image which can be used for anomaly detection. Results on surface images of decorated plastic parts demonstrate that this approach is suitable for detection of visible anomalies and moreover surpasses all other tested methods.		
	\end{abstract}
	

	%
	\IEEEpeerreviewmaketitle

	\section{Introduction}\label{sec: intro}
	%
	%
	%
	%
	\IEEEPARstart{S}{urface} inspection is an issue in many manufacturing industries. This is particular the case where the visual quality impression of a product has a strong influence on the decision whether a product is purchased or not. In these cases, it is common practice to inspect every fabricated part for visible defects, either manually or automatically. Manual inspection suffers from being a monotonous task, which typically leads to overlooked errors and subjective assessments. For these and other reasons, the industry's ambitions are high to automate any type of surface inspection. One particular challenge in automated surface inspection is the distinction of permitted structures from defects on patterned or textured surfaces. Another challenge are components which show slight but permitted sample-to-sample appearance variations. This fingerprint-like behavior means that reference samples ("golden samples") cannot be used directly to filter out the permitted appearances.
	
	Surface inspection can be linked to the field of anomaly and novelty detection, which is the detection of patterns that deviate from the expected behavior or the detection of unseen patterns \cite{Chandola.2009}. In those cases, a model is built from normal (fault-free) samples only. Anomalous (faulty) patterns are then detected by monitoring the anomaly score. Nowadays, convolutional neural networks (CNNs) are the predominant choice for many image related tasks in machine learning. Although they are typically trained in a supervised manner in tasks such as classification, object detection and segmentation, they can also be used for unsupervised tasks such as dimensionality reduction, unsupervised clustering \cite{Turchenko.2017} and image generation tasks such as the generation of completely novel images \cite{Dosovitskiy.2015} or image completion \cite{Pathak.2016, Iizuka.2017, Yu.2018}.
	
\begin{figure*}[!t]
		\centering
		\includegraphics[width=6.8in]{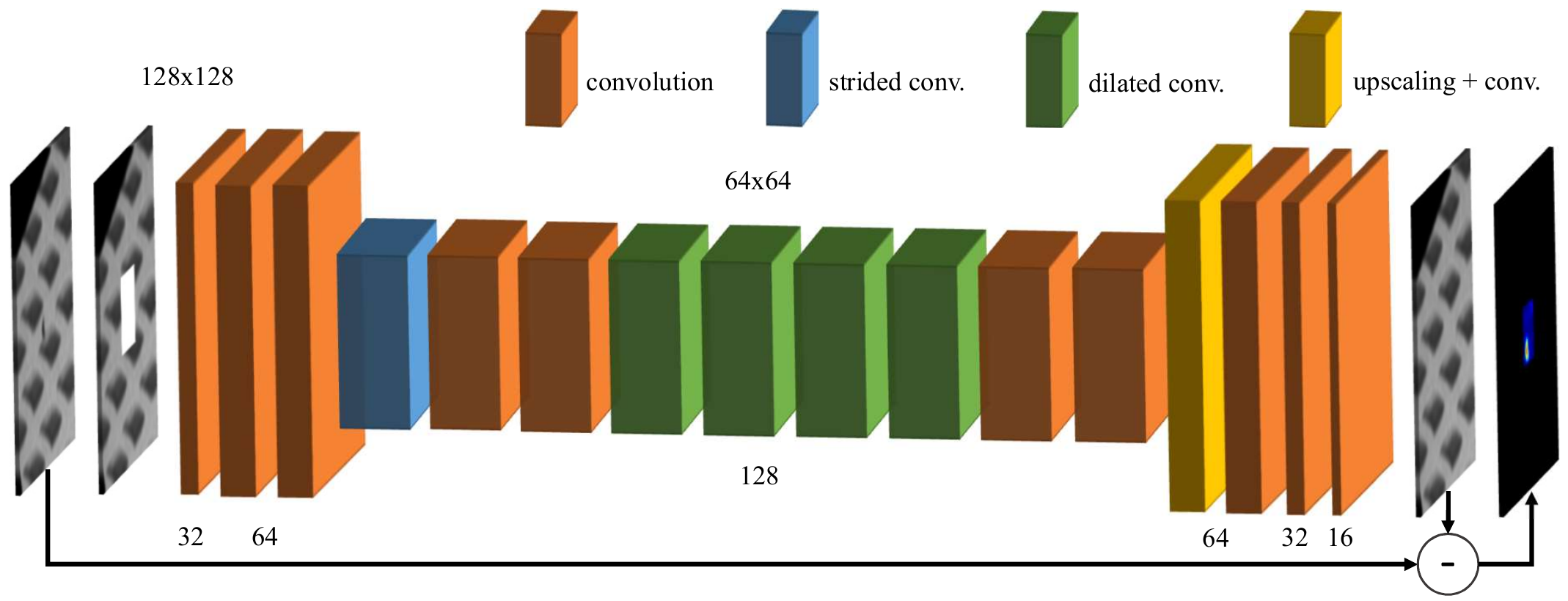}
		\caption{Schematic representation of the proposed algorithm. The $32\times32$ central region of the input image patch ($128\times128$) is removed (set to zero) before being fed into the deep convolutional neural network. After image reconstruction by the network the central region is subtracted from the corresponding query region. The absolute value of this difference image  is used as anomaly score. The image completion network consists of 17 layers: $\text{Conv}(5, 1, 1, 32) - \text{Conv}(3, 1, 1, 64) - \text{Conv}(3, 1, 1, 64) - \text{Conv}(3, 1, 2, 128) - \text{Conv}(3, 1, 1, 128) - \text{Conv}(3, 1, 1, 128) - \text{Conv}(3, 2, 1, 128) - \text{Conv}(3, 4, 1, 128) - \text{Conv}(3, 8, 1, 128) - \text{Conv}(3, 16, 1, 128) - \text{Conv}(3, 1, 1, 128) - \text{Conv}(3, 1, 1, 128) - \text{Bilinear Upscaling}( 2\times) - \text{Conv}(3, 1, 1, 64) - \text{Conv}(3, 1, 1, 64) - \text{Conv}(3, 1, 1, 32) - \text{Conv}(3, 1, 1, 16) - \text{Conv}(3, 1, 1, 1) - \text{Clip}(-1, 1)$, where $\text{Conv}(k, d, s, c)$ denotes a convolutional layer with kernel size $k\times k$, dilation rate $d$, stride $s$ and $c$ output channels.}
		\label{net}
	\end{figure*}

	In this paper, a deep convolutional neural network is used for patch-wise completion of surface images with the aim to detect aesthetic surface defects. Since the network is trained exclusively on normal data, it can be used to compute fault-free clones of the completed region. By subtracting the corresponding query region, a pixel-wise anomaly score is obtained, which is then used to detect defects (see Fig. \ref{net}).
	
	\begin{figure}[!t]
		\centering
		\includegraphics[width=3.5in]{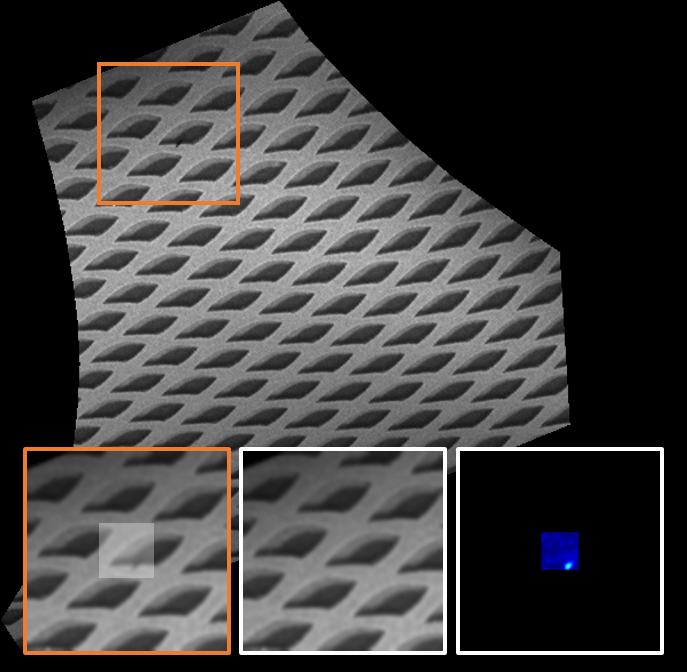}
		\caption{Examplary surface used as test case in this study. The three images below show the image patch fed to the proposed pipeline, the reconstructed fault-free clone and the resulting pixel-wise anomaly score.}
		\label{fig_sim}
	\end{figure}

	Image completion tasks typically have the aim to complete missing regions of an image in the most natural looking way. Besides being semantically meaningful, the inpaint must also look as authentic as possible. For this reason, feed-forward inpainting DCNNs are often trained jointly with an adversarial network, which was first done by Yu et al. in 2016 \cite{Pathak.2016}. The adversarial network has the objective to discriminate between fake and real images. In contrast, the generative model must increase the error rate of the adversarial network by generating realistic images. Although this additional adversarial loss indeed causes inpaints to look more realistic, it has no positive effect on pixel-wise matching the missing part of the image. Training with the joint loss function even increases the pixel-wise reconstruction error, which is undesirable behavior for anomaly detection. For this reason, in this paper the feed-forward generative DCNN is trained with reconstruction loss only. The reconstructed fault-free surface images of the case study are matched so closely that they are hardly distinguishable from the query images. It is shown that the detection of well recognizable defects on patterned surfaces is possible with this approach.
	
	
	
	\section{Related Work}
	Anomaly detection---closely related to outlier detection and novelty detection--- is of interest in many research areas and application areas. Hence, there exist several reviews on this issue \cite{Chandola.2009, Omar.2013, Pimentel.2014, Agrawal.2015, Kwon.2017}. There are also some reviews about surface inspection \cite{Xie.2008, Neogi.2014, Huang.2014}, which can be considered a special case of anomaly detection. Among the learning-based methods one can distinguish between two groups: One-class and multi-class learning. The multi-class settings include both normal (fault-free) and faulty samples during training. The advantage is the straight-forward application of supervised DCNN based image classification or segmentation architectures such as VGG \cite{Simonyan.2014b} or FCN \cite{JonathanLong.2015}. The disadvantage, however, is that a large quantity of faulty instances has to be collected. This is especially difficult in scenarios where defects occur very rarely. Moreover, manual labeling of defects is extremely laborious, especially at pixel level.
	
	One-class approaches, on the other hand, only require normal instances for training. In general, however, training works well even if a small percentage of the data is not fault-free. If only one-class data is available, typically supervised image classification and segmentation algorithms cannot be applied. A hybrid approach is the injection of artificial defects. It enables the application of classification and segmentation algorithms on fault-free raw data \cite{Haselmann.2017b}. However, the detection of defects can only be guaranteed for those that are represented by the distribution of synthetic defects. On the other hand, even tiny and weakly contrasted defects can be detected, provided that similar synthetic defects have been injected. 
	
	Learning based anomaly detection methods, that are purely trainable with one-class data, are for example one-class SVMs, which were first introduced by \citeauthor{Scholkopf.2001} \cite{Scholkopf.2001}. \citeauthor{Li.2014} \cite{Li.2014} used them in combination with clustering for outlier detection of face images. Another one-class anomaly detection method was proposed by \citeauthor{Zhang.2017} \cite{Zhang.2017}, who used a CNN to map normal instances into a certain feature space, in which the mapped instances were clustered within a hypersphere. While both of these approaches worked well for detecting images strongly deviating from the normal ones, weakly deviating images were often misclassified. Apparently they are therefore less suitable for the detection of small anomalies on surface images. Furthermore, no pixel-wise anomaly detection is provided.
		
	Another possibility to train on one-class data are reconstruction-based or image generating and completing approaches. These approaches---often based on neural networks---have the advantage that pixel-wise anomaly detection is possible. For example, autoencoders are used to reconstruct an input after passing it through a bottleneck layer \cite{Xu.2016} \cite{Kholief.2017}. However, they tend to compress image content without learning a semantically meaningful representation. Denoising variants address this problem by corrupting the input image where the DAN has the objective to reconstruct the uncorrupted query \cite{Mei.2018}. Since, however, this corruption usually is done on pixel level, the DAN is not required to learn a lot of semantical information \cite{Pathak.2016}. A very recent image generating approach is based on a GAN that is trained to generate normal instances. When inspecting an image for anomalies, the GAN is used within an iterative optimization process to generate an image that looks as similar as possible \cite{Schlegl.2017} to the inspected image. Images with anomalies will not be generated since they are not part of the training data. The approach, however, is very slow due to the optimization process. Therefore, the method is difficult to apply in scenarios such as surface inspection, where thousands of image patches have to be processed within seconds.
	
	Similarly to denoising autoencoders (DANs), image completion networks have to deal with corrupted images. However, instead of containing low-level corruptions, a whole region of the input image is missing. An image completion algorithm, therefore, has to semantically understand the image. Similarly to anomaly detecting GANs \cite{Schlegl.2017}, an image completion network can be used for anomaly detection. An inpainted region is expected to be fault-free whenever the training data consists only of normal samples. For example, \citeauthor{Munawar.2015} \cite{Munawar.2015} used Boltzmann machines to inpaint road images with the aim to detect anomalies such as obstacles.
	

	\section{Methodology}
	
	\subsection{Image Completion Network}
	
	\begin{figure}[!t]
		\centering
		\includegraphics[width=3.5in]{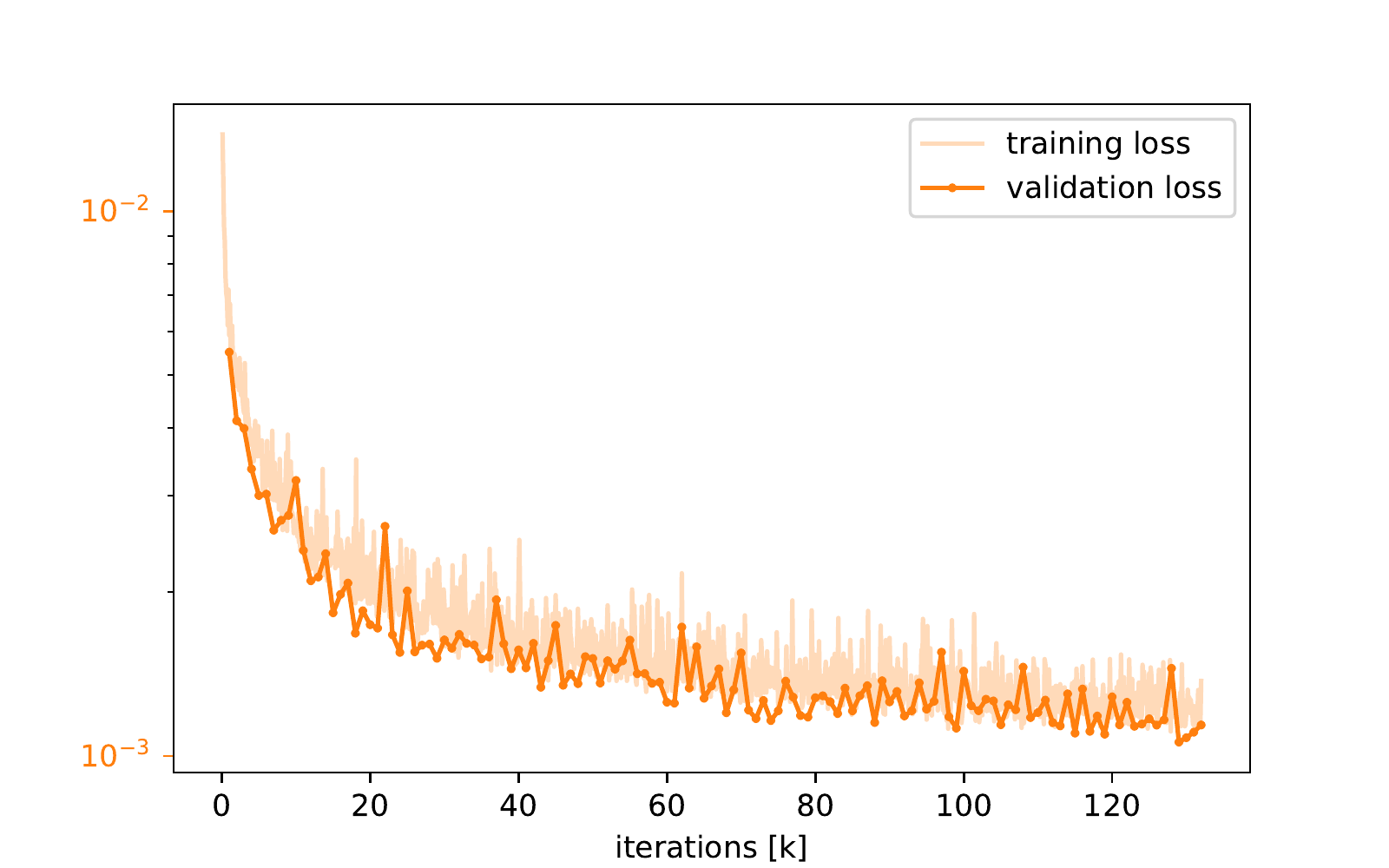}
		\caption{Training of the DCNN on data set A.}
		\label{training}
	\end{figure}
	
	\textbf{Architecture:} A fully convolutional network adapted from \cite{Iizuka.2017} and \cite{Yu.2018} is used for image completion. Overall, the network consists of 17 layers (see Fig. \ref{net}). After the third layer, the resolution of the feature maps is halved by strided convolution. In order to increase the receptive fields of the output neurons, a series of dilated convolutions according to \cite{Iizuka.2017} is used (layers 7 - 10). Upscaling back to the input size at layer 13 is performed by bilinear rescaling followed by a convolution. In accordance to \cite{Yu.2018}, mirror padding is used for all convolutional layers. Further, Exponential Linear Unit (ELU) activation functions are used.
	
	\textbf{Loss function:} The network is trained with L1 reconstruction loss. The $32\times32$ center region, defined by the binary mask $M$, is weighted differently from the remaining region. With $X$ being the image patch to be inspected, the network is trained with the loss function
	
	
	\begin{alignat*}{3}
	\mathcal{L}(X) & =  &\lambda&  &\left \Vert 
	M \odot \left(X-F(\bar{M}\odot X)\right)
	\right \Vert_{1} /{N} &\\
	&+ 
	(1- & \lambda&) &\left \Vert 
	\bar{M}\odot \left(X-F(\bar{M}\odot X)\right)
	\right \Vert_{1} /{N}&
	\end{alignat*}
	
	\begin{figure*}[!t]
		\centering
		\subfloat[Precision-Recall (PR) curve]{\includegraphics[width=3.0in]{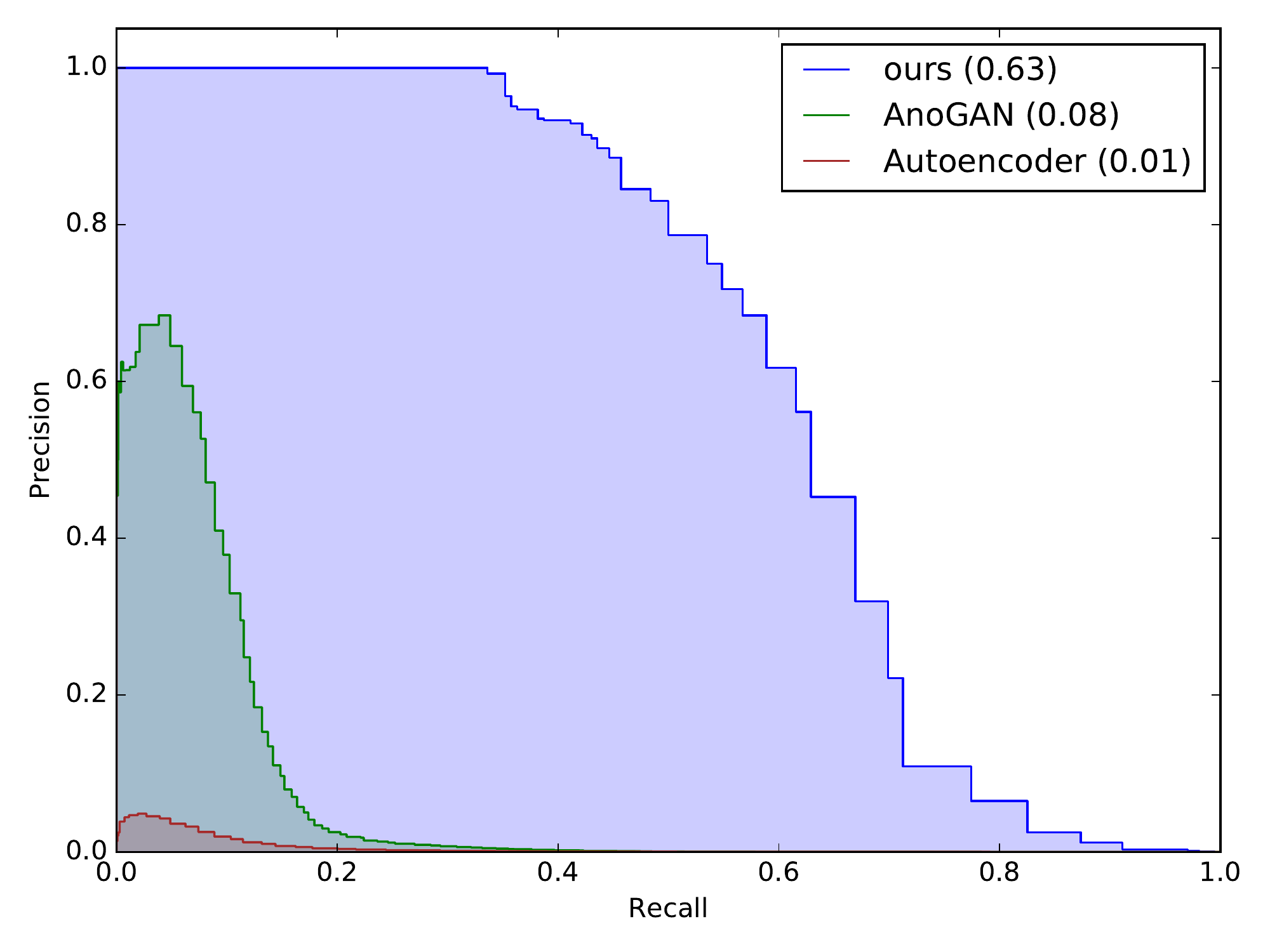}%
			\label{Pixelwise scores}}
		\hfil
		\subfloat[Receiver Operator Characteristic (ROC) curve]{\includegraphics[width=3.0in]{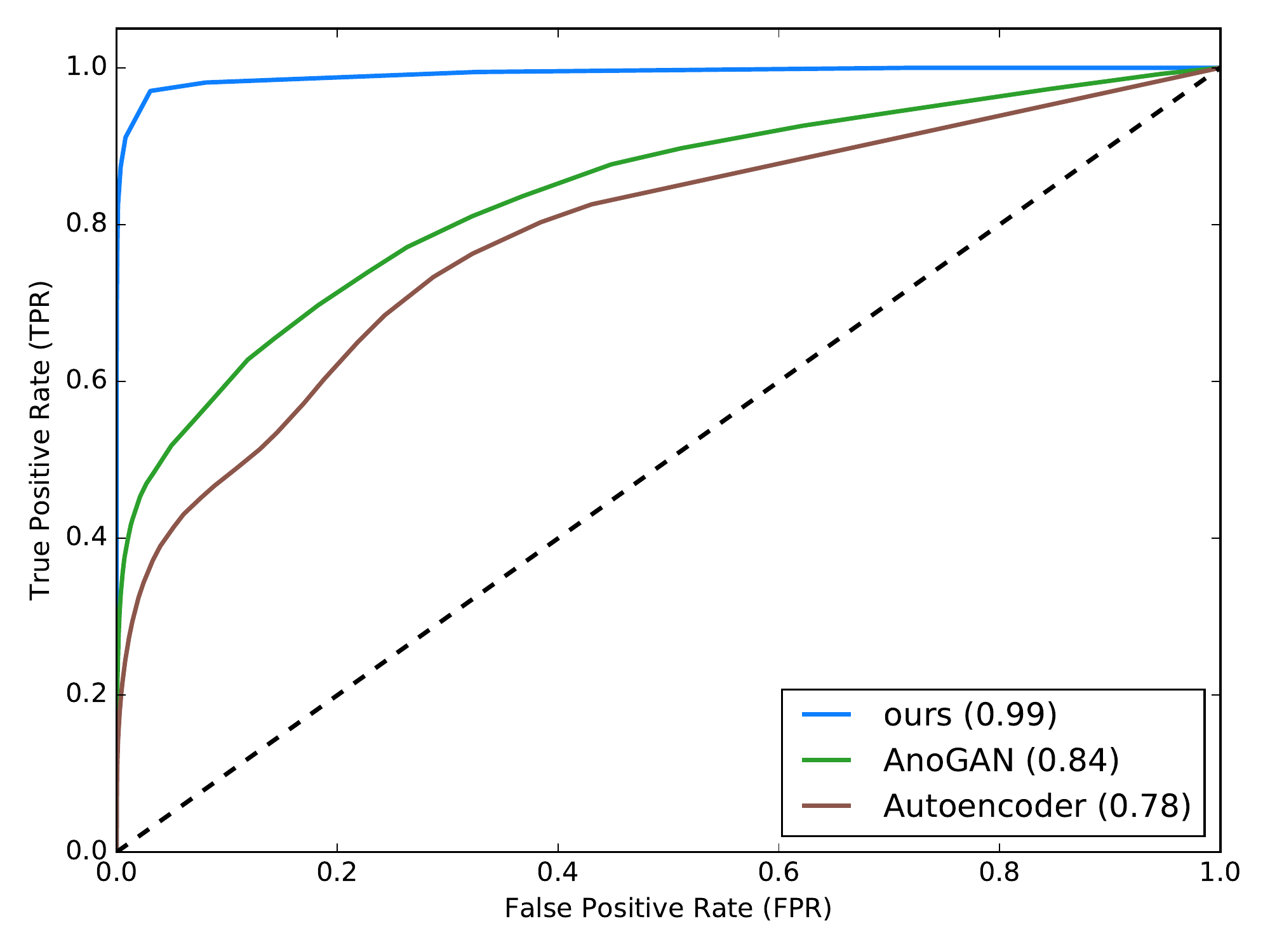}%
			\label{fig_second_case}}
		\caption{PR and ROC curve of all tested methods on data set A.}
		\label{prc}
	\end{figure*}

	where $\odot$ denotes element-wise matrix multiplication, $\bar{M}$ denotes the complement mask of $M$, $\lambda$ is the weighting parameter between the center and the remaining region, $N$ is the number of pixels of $X$, and $\Vert \cdot \Vert_{1}$ denotes the L1 matrix norm. The parameter $\lambda$ was chosen to be 0.9 for the conducted experiments. Since only the reconstruction error is relevant for anomaly detection, an adversarial loss is not included. Only the pixel-wise reconstruction error is relevant for anomaly detection. Hence, an adversarial loss is not included, since a more natural looking image does not decrease the reconstruction error.
	
	\textbf{Training details:} The image completion network was trained from scratch using the ADAM optimizer \cite{Kingma.2015} with hyper parameters $\alpha= 0.0002$, $\beta_1 = 0.9$, $\beta_2 = 0.999$, $\epsilon = 10^{-8}$ and a batch size of 128. All weights were initialized from a truncated Gaussian distribution with a mean of 0 and a standard deviation of 1. The biases were zero-initialized. Each model was trained on 100k-200k batches within 24 hours on a GTX 1080Ti (see Fig. \ref{training}).

	\subsection{Pre- and post processing}
	The image completion network is fed with image patches of size $128\times128$. The patches are corrupted by masking out the central area of size $32\times32$. This large ratio between known and unknown image content provides the network with more semantic information to complete the center region. 
	
	Image patches are extracted in real-time from high resolution surface images on randomized positions. This way, all possible patches are extracted from the raw data during training. In addition, data augmentation consisting of randomized image transformations is performed. This, however, usually leads to unwanted border effects. In order to avoid these effects, patches larger than the target size are extracted before applying these transformations and center-cropped to size $128\times128$ afterwards.

	After reconstruction of the corrupted image by the network, the pixel-wise absolute difference between the reconstructed image and the query image is computed. For anomaly detection, only the $24\times24$ center region of this absolute difference image is used. Image patches in which defects appear close to the border of the cut out $32\times32$ center region, the neural network seems to generate local continuations of the bordering defects. By considering only the $24\times24$ center region, these unwanted continuations are mostly excluded. In order to ensure that anomalies in the test data end up in this window as a whole at least once, a moving window with stride 16 is used.
	
	\subsection{Reported metrics}
	In contrast to the training and validation data, which is fault-free for the most part, the test data set contain significantly more defects including highly visible ones. Nevertheless, the test data set is extremely class imbalanced, especially on pixel level. Certain metrics, such as accuracy, are therefore misleading. For this reason the precision-recall curves (PRCs) are reported. Additionally the more commonly used ROC curves are reported. All metrics are computed on pixel level.
	
	\section{Experiments}
	
	\subsection{Data sets}
	The experiments are conducted on surface images of decorated free-form plastic parts. The part's decoration show significant part-to-part variations in terms of distortion, location and appearance of the pattern primitive. The part positioning, however, remains constant from part to part. Multiple views are provided per part in order to cover the whole surface to be inspected. Overall, two types of decorated parts are tested, resulting in two data sets (A and B). In data set A, 4 views are provided per part. Data set A comprises 34 parts for training, 8 for validation and 17 for testing. The test set includes several different visible defects that have been manually labeled. In data set B, there are 3 views per part with 12 parts for training, 12 for validation and 4 for testing. Both data sets include very eye-catching defects as well as very low contrasted defects that are hardly visible for humans.
	
	\subsection{Results}
	
	\begin{figure*}[!t]
		\centering
		\subfloat[Clearly visible defects on data set A.]{\includegraphics[width=3.in]{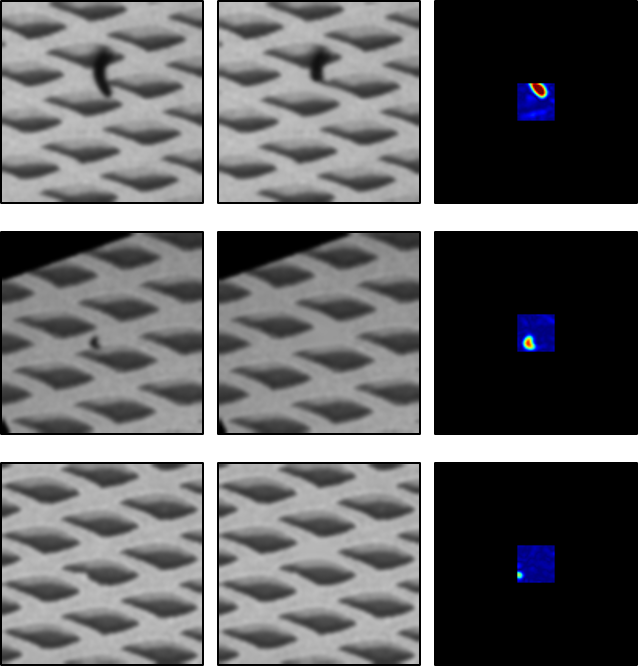}%
			\label{shark_intensive_defects}}
		\hfil
		\subfloat[Weakly recognisable defects on data set A.]{\includegraphics[width=3.in]{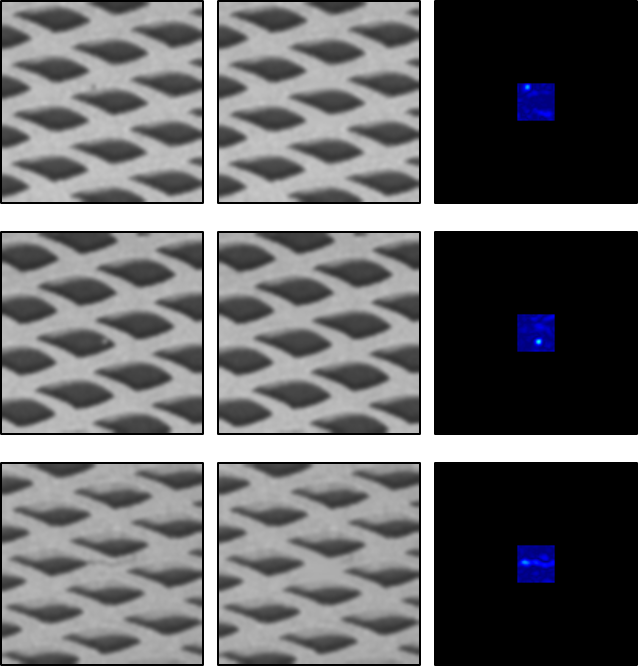}%
			\label{shark_medium_defects}}
		\hfil
		\subfloat[Defects on data set A that cannot be detected.]{\includegraphics[width=3.in]{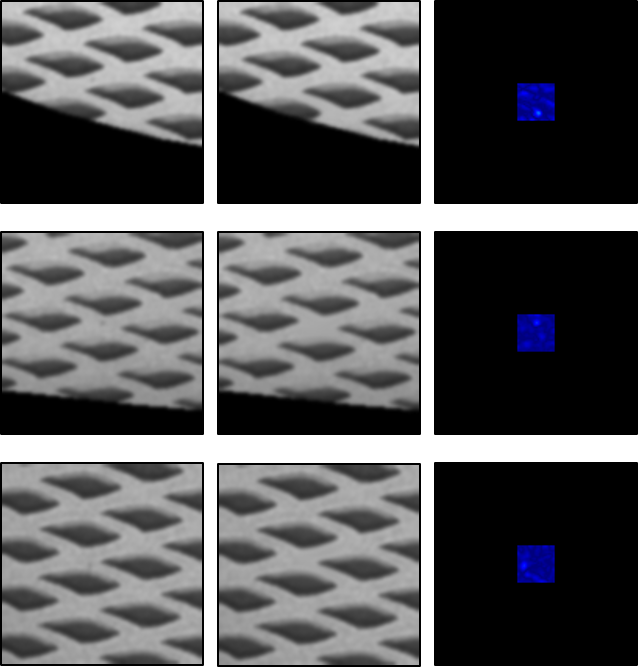}%
			\label{shark_2weak_defects}}
		\hfil
		\subfloat[Defects on data set B.]{\includegraphics[width=3.in]{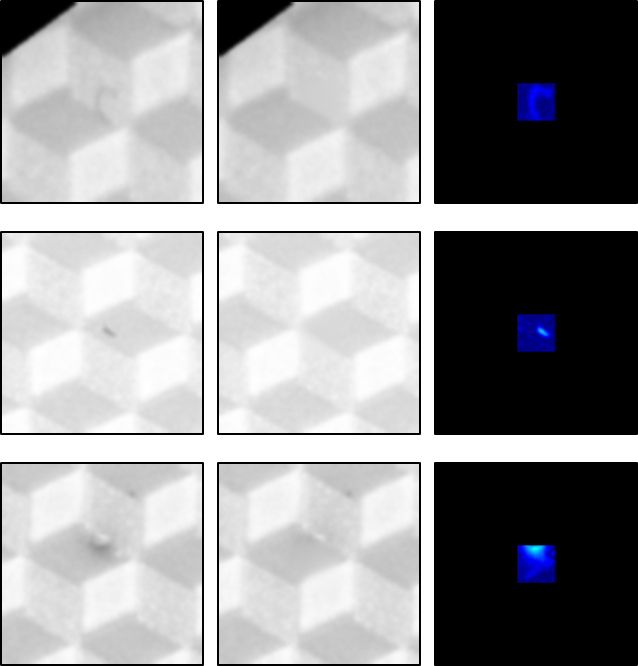}%
			\label{cubic_defects}}
		\caption{Examples of image patches that are put through the proposed pipeline. Each row (consisting of three images) of each sub figure shows one example. The images on the left show the query images. The images in the middle show the images reconstructed by the DCNN. Note that the center regions of the reconstructed images do not contain the defective structures of the query images. The images on the right show the pixel-wise absolute values of the difference images between query images and reconstructed images for the corresponding center regions. Fig. \ref{shark_intensive_defects} and Fig. \ref{shark_medium_defects} show examples of defects that can be detected by the proposed method. Fig. \ref{shark_2weak_defects} shows examples of defects that are too weakly contrasted to be detected. Fig. \ref{cubic_defects} shows examples of data set B.}
		\label{example_datasetA}
	\end{figure*}
	
	The proposed algorithm is compared to a classical autoencoder (bilinear downscaling to size $32\times32$, followed by $\text{FC}(32^2, 128) - \text{ReLU} - \text{FC}(128, 32^2)$ and upscaling back to size $128\times128$) and AnoGANs. Latter were recently introduced in 2017 by \citeauthor{Schlegl.2017} \cite{Schlegl.2017}. For the comparison with AnoGANs, the implementation of \citeauthor{Ayad.2017} \cite{Ayad.2017} was applied on patches of size $64\times64$. Since AnoGANs require a time intensive optimization process to generate one single image patch, comparisons are only performed on data set A. AnoGANs seem to fail to reconstruct image patches that show visible edges of the region of interest. For this reason, such patches are not considered in the reported metrics. Furthermore, for the evaluation of AnoGANs, image patches that were obviously reconstructed incorrect (presumably due to the optimizer being stuck in a local minimum) are also not considered in the reported metrics.

	On data set A, the proposed algorithm achieves an AUROC (area under ROC curve) of 0.99 and an AUPRC (area under PR curve) of 0.63. Clearly visible anomalies could be reliably detected (see Fig. \ref{shark_intensive_defects}a). Most of the weakly contrasted defects also stand out in the anomaly image compared to residual patterns, which are a result of slight mismatches between generated and query image (see Fig. \ref{shark_medium_defects}b). As suspected, however, very weakly contrasted defects (see Fig. \ref{shark_2weak_defects}c) cannot be distinguished from the residual pattern structures in the anomaly image. Nevertheless, the proposed algorithm clearly surpasses the other tested methods in both AUROC and AUPRC (see \ref{table_example}). On data set B, the model was also trained from scratch, achieving a similarly high AUROC (0.98) and AUPRC (0.26) (see Fig. \ref{cubic_defects}d).
		
		
		\begin{table}[]
			\renewcommand{\arraystretch}{1.3}
			\caption{Summary of results}
			\label{table_example}
			\centering
			\begin{tabular}{r|c|c}
				\multicolumn{1}{l|}{} & AUPRC & AUROC \\ \hline
				ours (A) & 0.63 & 0.99 \\
				AnoGAN (A) & 0.08 & 0.84 \\
				Autoencoder (A) & 0.01 & 0.78 \\
				ours (B) & 0.26 & 0.98 \\ \hline
			\end{tabular}
		\end{table}
		
		\section{Conclusion}
		In this paper, the evaluation of surface inspection data is performed by a deep convolutional neural network that is trained to complete image patches whose center region is cut out. Since the network is trained exclusively on fault-free data, it completes the image patches with a fault-free version of the missing image region. The reconstruction error provides a pixel-wise anomaly score, which is subsequently used for defect detection. The network is trained with L1 reconstruction loss with the missing center region being weighted differently compared to the remaining region. Despite not utilizing an adversarial loss, the generated image patches can barely be distinguished from real ones. Distinctly contrasted anomalies can be detected, while very weakly contrasted ones are often confused with residual patterns. On data set A, the proposed algorithm achieves a pixel-wise AUROC (area under ROC curve) of 0.95 and a pixel-wise AUPRC (area under PRC curve) of 0.43 and thus clearly surpasses the other tested methods.

		\section*{Acknowledgment}
		The research work of this paper was performed at the Polymer Competence Center Leoben GmbH (PCCL, Austria) within the framework of the COMET-program of the Federal Ministry for Transport, Innovation and Technology and the Federal Ministry of Economy, Family and Youth with contributions by BurgDesign GmbH, HTP GmbH and Schoefer GmbH. The PCCL is funded by the Austrian Government and the State Governments of Styria and Upper Austria.

		\ifCLASSOPTIONcaptionsoff
		\newpage
		\fi

		
		
		\bibliographystyle{IEEEtranN}
		\bibliography{IEEEabrv}
	\end{document}